\documentclass[final]{cvpr}

\usepackage{times}
\usepackage{epsfig}
\usepackage{graphicx}
\usepackage{amsmath}
\usepackage{amssymb}

\usepackage{graphicx}
\usepackage{comment}
\usepackage{amsmath,amssymb} 
\usepackage{color}
\usepackage{array}
\usepackage[table,xcdraw]{xcolor}
\usepackage{multirow}
\usepackage{wrapfig}
\usepackage{lipsum}
\usepackage[pagebackref=true,breaklinks=true,colorlinks,bookmarks=false]{hyperref}

\def\cvprPaperID{3} 
\def\confYear{CVPR 2021}

\begin{document}

\title{Graph-based Person Signature for Person Re-Identifications}
\author{Binh X. Nguyen$^{1}$, Binh D. Nguyen$^{1}$, Tuong Do$^{1}$, Erman Tjiputra$^{1}$, Quang D. Tran$^{1}$, Anh Nguyen$^{2}$\\
{$^{1}$AIOZ, Singapore}\\
{$^{2}$Imperial College London, UK}\\
{\tt\small \{binh.xuan.nguyen,binh.duc.nguyen,tuong.khanh-long.do,erman.tjiputra,quang.tran\}@aioz.io}\\
{\tt\small a.nguyen@imperial.ac.uk}}

\maketitle

\begin{abstract}
The task of person re-identification (ReID) is to match images of the same person over multiple non-overlapping camera views. Due to the variations in visual factors, previous works have investigated how the person identity, body parts, and attributes benefit the person ReID problem. However, the correlations between attributes, body parts, and within each attribute are not fully utilized. 
In this paper, we propose a new method to effectively aggregate detailed person descriptions (attributes labels) and visual features (body parts and global features) into a graph, namely Graph-based Person Signature, and utilize Graph Convolutional Networks to learn the topological structure of the visual signature of a person. The graph is integrated into a multi-branch multi-task framework for person re-identification. The extensive experiments are conducted to demonstrate the effectiveness of our proposed approach on two large-scale datasets, including Market-1501 and DukeMTMC-ReID. Our approach achieves competitive results among the state of the art and outperforms other attribute-based or mask-guided methods. Source available at \small\url{https://github.com/aioz-ai/CVPRW21_GPS}.
\end{abstract}

\section{Introduction}
\label{sec:intro}
Person re-identification (ReID) aims to retrieve a particular person image in a collection of images captured by multiple cameras from various viewpoints across time. The challenges of the person ReID task come from significant variations of human attributes such as poses, gaits, clothes, as well as challenging environmental settings like illumination, complex background, and occlusions. With the rise of deep learning, most of the recent studies utilize Convolutional Neural Network (CNN) to tackle the person ReID problem. Many approaches have been proposed such as metric learning~\cite{triplet, nguyen2020deep,chen2017beyond}, attention-based~\cite{mancs,sona,scal, nguyen2019v2cnet}, GAN-based~\cite{PN-GAN,FD-GAN,dg-net}, attribute-based \cite{apdr,attreid,a3m,mlfn,affnet,PAAN}, and spatial-temporal-based methods~\cite{st-ReID}.

Recently, attribute-based methods have shown great success in providing semantic features for the deep network~\cite{affnet,aanet}. Unlike the person identity label, which offers only coarse information to identify one identity among all other person identities, the attributes are the detailed descriptions that are highly intuitive and mostly unchanged between images captured from different cameras. Therefore, they can be used to explicitly guide the model to learn a robust person representation by defining human characteristics. Furthermore, as shown in~\cite{attreid}, attributes can also be used to speed up the retrieval process of the person ReID task by filtering out images from the gallery that do not share the same attributes with the probe image.

\begin{figure*}
    \centering
    \includegraphics[width=0.7\textwidth, keepaspectratio=true]{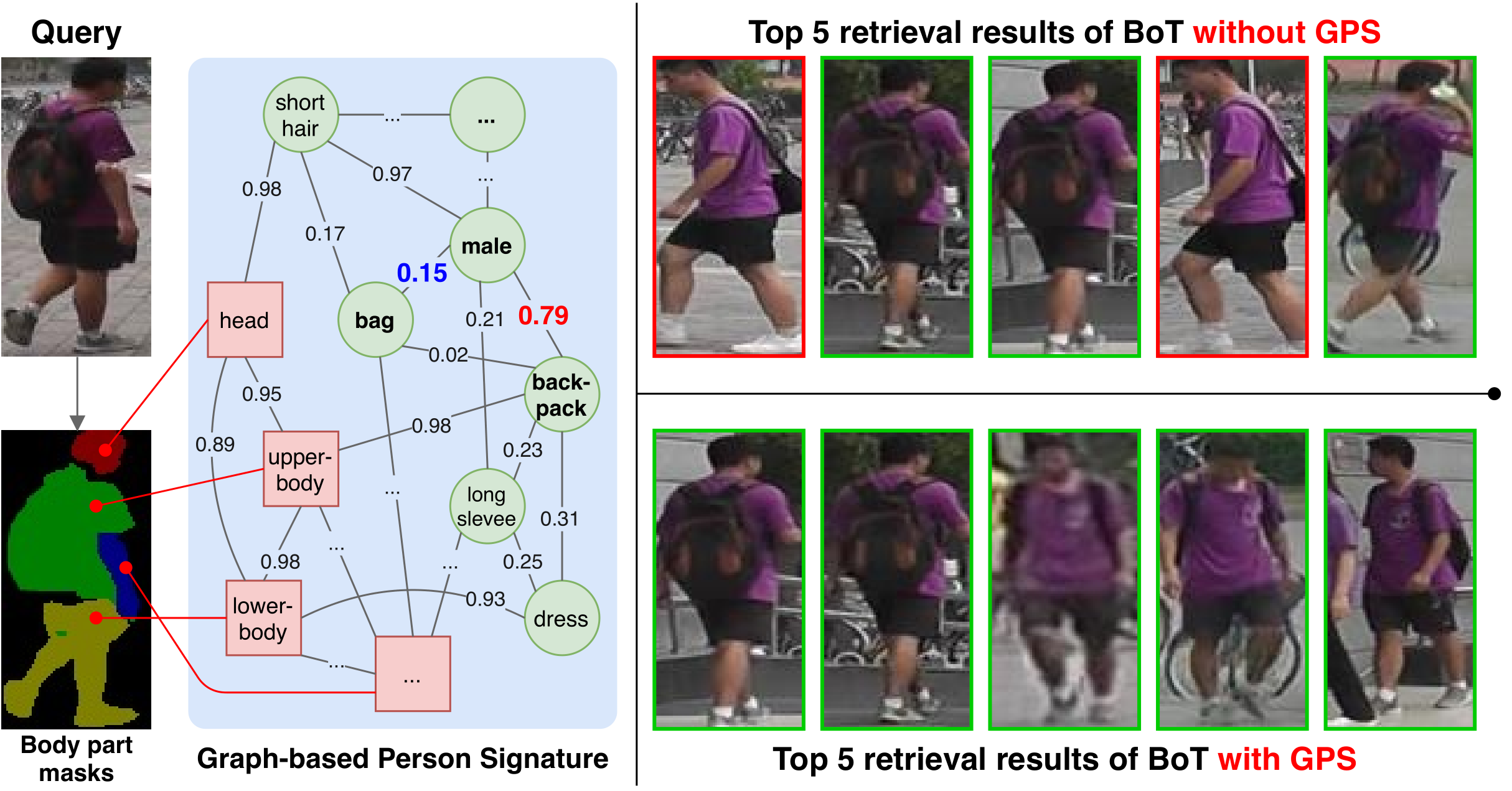}
    \caption{The effectiveness of our GPS in improving retrieval results on Market-1501 dataset \cite{market1501}. Note that the green/red boxes denote true/false retrieval results, respectively. The retrieval results of the baseline model BoT \cite{BoT} have some different attributes, e.g., `bag' and `backpack'. This leads to false retrieval results at rank-1 and rank-4. However, when integrating our GPS into BoT, these false results are removed. Our GPS gives the correlations between attributes and body parts (the graph in blue background). The correlation between `male' and `backpack' is higher than between `male' and `bag' (i.e., 0.79 $>$ 0.15). Therefore, the information extracted from GPS makes the person feature more discriminative and consequently improves the results.}
    \label{fig:vis_intro}
\end{figure*}

In this work, we propose to utilize the person attribute information with its associated body part to encode the visual person signature in one unified framework. We hypothesize that the detailed person descriptions (attributes labels) can be integrated with visual features (body parts and global features) to create a unique signature for a particular person. Since both body parts and attributes provide local representations, by linking them together, the network can have a better understanding of the relationship between visual features and attribute descriptions. Although previous works have investigated how person identity, body parts, and attributes benefit the task of person ReID~\cite{attreid,PAAN,ACRN,aanet}, our key difference is that we utilize Graph Convolutional Networks (GCN) to effectively construct and model the correlation between attributes and body parts with global features. In particular, we treat body part regions and attributes as nodes in a graph and utilize a GCN to learn the topological structure of a person's signatures. The GCN propagates messages on a graph structure. After message traversal on the graph, the node's final representations are obtained from its data and from other node's information. Fig.~\ref{fig:vis_intro} shows the effectiveness of our approach.

\section{Related work}
\label{sec:related_work}
Methods based on deep convolutional networks have dominated in the ReID community. In this section, we thoroughly review two popular approaches similar to our proposed method: Part-based approach and Attribute-based approach. Other approaches are briefly also mentioned.

\textbf{Part-based approach.}
Several efforts \cite{mgn,LocalCNN,dsa} learn meaningful features from local parts of a probe image to improve the representativeness of a person by incorporating the body part region information, i.e., the local spatial information. In particular, the local CNN method \cite{LocalCNN} proposed to split the backbone network into multiple parts horizontally and incorporate the local region information at each layer. Meanwhile, MGN \cite{mgn} used a multi-branch network; each branch takes responsibility for extracting coarse-to-fine information. In \cite{p2net}, the authors proposed to accurately combine human parts and the coarse non-human parts with a self-attention mechanism. SPReID \cite{SPReID} was proposed to integrate semantic human parsing to person ReID with local visual cues of human parts. Song et al. \cite{MGCAM} combined binary segmentation mask with a mask-guided attention model for the person ReID task.

\textbf{Attribute-based approach.} Many studies utilize attribute information to improve the representation of local features in the person ReID task. In \cite{attreid}, the authors proposed an attribute-person recognition (APR) network which learns person ReID and person attribute recognition simultaneously. $A^3M$ \cite{a3m} introduced a new attention mechanism by learning attribute-guide attention and category-guided attention reciprocally. In \cite{ling2019improving}, the authors proposed a multi-task learning framework with four subtasks for person ReID. In \cite{affnet}, the authors proposed AFFNet, which is a multi-branch model that fuses features from person identity branch and attribute branch. In \cite{PAAN}, the authors proposed a multi-branch model levering both identity label and attribute information. The AANet \cite{aanet} combined the global representation with three tasks, including person ReID, body part localization, and person attribute recognition. The APDR \cite{apdr} method fused attribute features and body part features to result in the final local features which are then concatenated with the global features for person re-identification.

\textbf{Other approaches.} Deep CNN has been used in various tasks in combining vision, language, and scene attributes \cite{sona, nguyennav,nguyen2019v2cnet, chen2020salience, do2018affordancenet,  nguyen2019object}. In \cite{pyramid}, the authors proposed a pyramid model that can match images at different scales by incorporating local and global information and the gradual cues between them. Considering the distance part-to-part relationship, in \cite{sona}, the authors proposed an attention mechanism to capture non-local and local correlations directly via second-order feature statistics. Inspired by GAN, Zheng et al. \cite{dg-net} proposed a joint learning framework that couples ReID learning and data generation in an end-to-end manner. More recently, in \cite{st-ReID} the authors proposed a two-stream spatial-temporal person ReID (st-ReID) framework that uses both visual semantic information and spatial-temporal information from the camera setting, thus eliminates lots of appearance ambiguity images.
Zhou et al.\cite{zhou2020online} proposed a online joint multi-metric adaptation algorithm which not only takes individual characteristics of testing samples into consideration but also fully utilizes the visual similarity relationships among both query and gallery samples. In \cite{chen2020salience}, the authors proposed the Salience-guided Cascaded Suppression Network which enables the model to mine diverse salient features and integrate these features into the final representation by a cascaded manner. In \cite{yang2020spatial}, Yang et al. proposed a Spatial-Temporal Graph Convolutional Network which enables to extract robust spatial-temporal information that is complementary with appearance information for video-based Person Re-identification task. The UnityStyle \cite{liu2020unity} method was proposed to smooth the style disparities within the same camera and across different cameras. Zhang et al. \cite{zhang2020relation} proposed the Relation-Aware Global Attention module which captures the global structural information for better attention learning. Besides, several methods are proposed to solve the problems of Occluded Person Re-Identification \cite{fpr,miao2019pose,gao2020pose,wang2020high}.


\section{Methodology}
\label{sec:method}
\begin{figure*}
    \centering
    \includegraphics[width=0.8\textwidth, keepaspectratio=true]{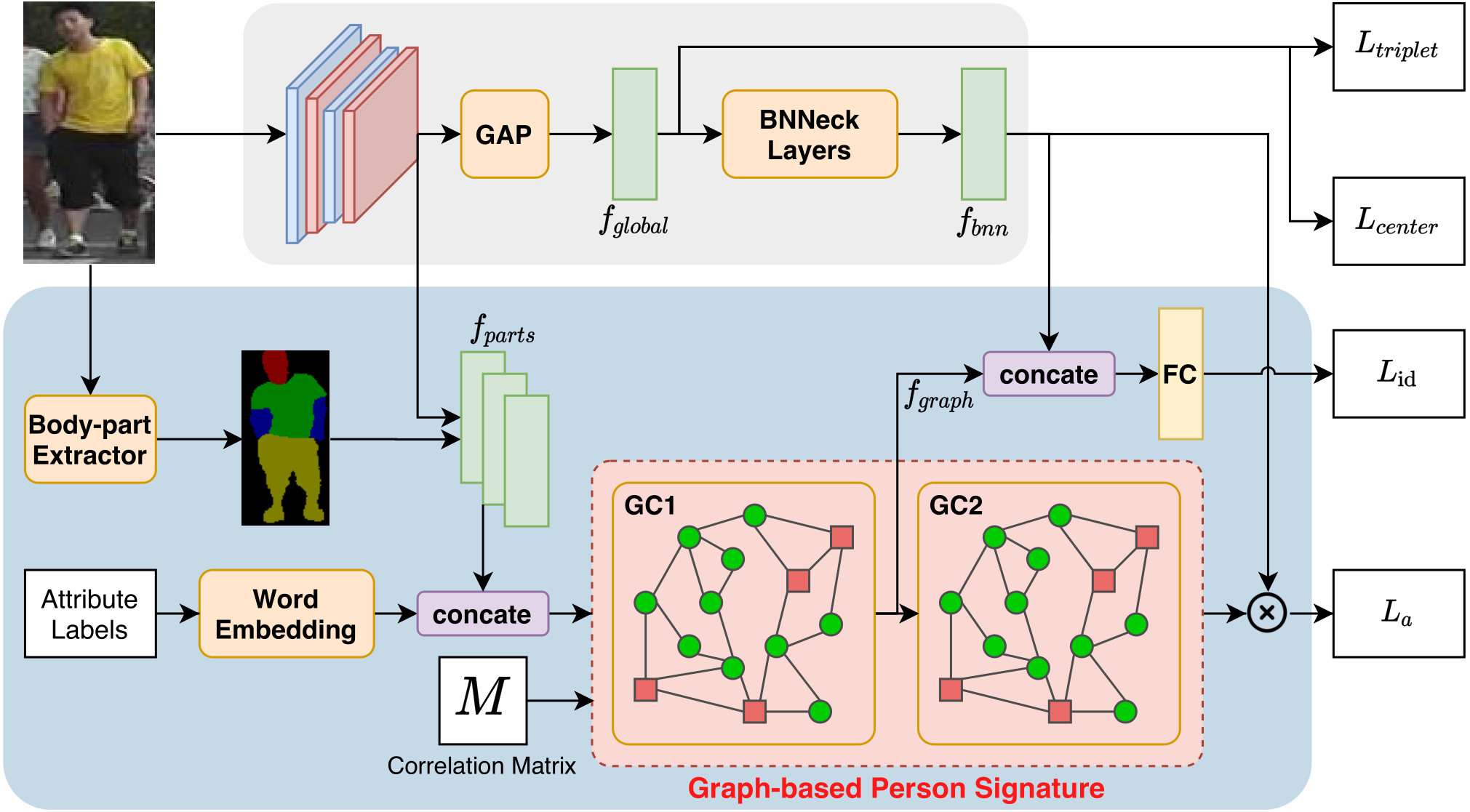}
    \caption{Illustration of our proposed framework including two branches: (1) global branch which extracts person global features; (2) GPS branch which performs reasoning the person attributes and body parts using GCN.} 
    \label{fig:framework}
\end{figure*}

The proposed framework is presented in Figure \ref{fig:framework}. We denote $I$ is a probe person image. This probe image $I$ is first passed through a backbone CNN to get the feature map $\mathbf{F}$. By utilizing a human parsing pretrained model, we extract the body part masks to obtain the visual features of each part. The person attributes are then represented by a lookup word embedding. Given body part features and attribute features, we construct the Graph-based Person Signature which includes attribute nodes and body part nodes conditioned on the correlation matrix. We employ the GCN \cite{kipf2016semi} for reasoning on the person signature graph and encoding the graph into more representativeness features. Our proposed method is a multi-branch multi-task framework for person ReID, where the main branch performs the verification task by optimizing two well-known loss functions: Triplet loss \cite{weinberger2006distance} and Center loss \cite{centerloss}. The auxiliary branch performs reasoning on the proposed person signature graph and solves the attribute recognition as well as the person identity classification tasks. The training process is explained in detail in Section \ref{sec:train}.

\subsection{Graph-based Person Signature: Construction}
\label{sec:graph}
The proposed GPS, denoted by $\mathcal{G} = (\mathcal{V},\mathcal{E})$, consists of nodes $\mathcal{V} = \{v_1, v_2, ..., v_N\}$ with the total number of nodes $N_G = N_A + N_P$, where $N_A$ is the number of attributes and $N_P$ is the number of body parts. Each node denotes either a person attribute or a human body part and is initialized with a $D_w$-dims feature vector $x_v$. The graph is represented by an adjacency matrix $\mathbf{M} \in \mathbb{R}^{N_G \times N_G}$ containing weights associated with each edge $(v_i, v_j) \in \mathcal{E}$. The correlation matrix $\mathbf{M}$ has the following form

$$\mathbf{M} = \left[\begin{matrix}
            \mathbf{AA}&\mathbf{AP}\\
            \mathbf{PA}&\mathbf{PP}
            \end{matrix}
    \right],$$
where $\mathbf{AA} \in \mathbb{R}^{N_A \times N_A}$ is the attribute-attribute correlation matrix, $\mathbf{PP} \in \mathbb{R}^{N_P \times N_P}$ is the parts-parts correlation matrix, $\mathbf{PA} \in \mathbb{R}^{N_P \times N_A}$ is the parts-attributes correlation matrix, and $\mathbf{AP} \in \mathbb{R}^{N_A \times N_P}$ is the attribute-parts correlation matrix.

\textbf{The attributes-attributes matrix.} We follow the process as described in \cite{chen2019} to construct the attributes-attributes matrix $\mathbf{AA}$. The element $\mathbf{AA}_{ij}$ denotes the probability of occurrence of attribute $j$ when the attribute $i$ occurs, which is formulated as follow
\begin{equation}
    \textbf{AA}_{ij} = \frac{L_{ij}}{K_i},
\end{equation}
where $K_i$ denotes the occurrence times of attribute $i$ in the training set, and $L_{ij}$ denotes the co-occurence of attribute pair $i$ and $j$.

\textbf{The parts-parts matrix.} We assume that the body parts are always recognizable for every probe image in the training set. Thus we set all elements in the $\mathbf{PP}$ matrix to 1. This can be inferred as if a body part $i$ is recognized, the probability of recognizing the body part $j$ is 1.

\textbf{The parts-attributes matrix.} The element of the matrix $\mathbf{PA}_{ij}$ denotes the probability of the attribute $i$ occurs when the body part $j$ is recognized. We establish a heuristic observation that some attributes only attached to a specific body, e.g., `hair length' is only attached to `head', not `lower body'. The body parts and their associate attributes are summarized in Table \ref{table:part-attr}. Based on the recognizable body parts assumption above, given body part $i$, $\mathbf{PA}_{ij} = 0$ if the attribute $j$ is not attached to the body part $i$, otherwise $\mathbf{PA}_{ij}=k_j$ where $k_j$ is the percentage of attribute $j$ occurs in the dataset.
\begin{table*}[!t]
\caption{Body parts and their associated attributes.}
\label{table:part-attr}
\centering
\begin{tabular}{l|l} 
\hline
\textbf{Body part} & \textbf{Attribute}                                                      \\ 
\hline
Foreground         & gender, age                                                             \\ 
\hline
Head               & hair length, wearing hat~                                               \\ 
\hline
Upper body         & upper clothing's type, upper clothing's color, carrying backpack~~      \\ 
\hline
Lower body         & lower clothing's type, lower clothing's color, lower clothing's length  \\ 
\hline
Arm                & sleeve length, carrying bag, carrying handbag~                          \\
\hline
\end{tabular}
\end{table*}

\textbf{The attributes-parts matrix.} The element of the matrix $\mathbf{AP}_{ij}$ denotes the probability of the body part $i$ is recognized while the attribute $j$ occurs. Due to our assumption that all body parts are recognizable in the dataset. $\mathbf{AP}_{ij} = 1$ if attribute $j$ is attached to body part $i$, otherwise, $\mathbf{AP}_{ij} = 0$.

In practice, the attributes are represented by word embedding $\mathbf{Z} \in \mathbb{R}^{N_A \times D_w}$, where $N_A$ is the number of attributes and $D_w$ is the dimensionality of word-embedding vector.

\textbf{Body parts representation.} To obtain the visual presentation of person body part, we utilize the state-of-the-art SCHP pretrained model \cite{li2019self} trained on LIP dataset \cite{lip} to predict the body part masks for all the images in advance. Although LIP dataset has 20 labels, in our work, we combine the labels to form the more coarse body part regions, namely \textit{head, upper, lower, arm, and foreground}. Note that the \textit{foreground} is the combination of other body parts, which represents the global attributes of a person such as \textit{age} and \textit{gender}. The parser segments each probe image into $N_P$ body parts represented by a set of masks $\mathcal{H} = \{\mathbf{H}_k\}_{k=1}^{N_P}$, where $\mathbf{H}_k$ is a binary mask with the same size as the probe image. 
Each mask $\mathbf{H}_k$ is scaled to the same size as feature map $\mathbf{F}$ and is applied $L_1$ normalization, which results in $\mathbf{H}'_k$. The feature map $\mathbf{F} \in \mathbb{R}^{W\times H \times D}$ has $W \times H$ locations, each location $i$ is associated with a feature vector $\mathbf{f}_i \in \mathbb{R}^D$. The $k$-th body part $f_{part}^{(k)}$ is computed as below:

\begin{equation}
    f_{part}^{(k)} = \sum\limits_{i=1}^{N_P} h_i^{(k)} \mathbf{f}_i,
\end{equation}
where $h_i^{(k)}$ is the scalar value at the location $i$ of $\mathbf{H}'_k$. The $ f_{part}^{(k)}$ is projected to a $D_w$-dim vector.

\subsection{Graph-based Person Signature: Reasoning}
Generally, GCN defines a multi-layer propagation process on a graph $\mathcal{G}$. Precisely, each layer in GCN is formulated as a function $f(\mathbf{X},\mathbf{M})$ which updates the node representations by propagating information between the input nodes $\mathbf{X} \in \mathbb{R}^{N_G \times D_w}$, where each row represents a node, under the guidance of correlation matrix $\mathbf{M}$.
Denoting $\mathbf{H}^{(k)}$ is the feature matrix after passing the input nodes $\mathbf{X}$ to $k$-th GCN layers. We follow GCN formulation proposed in \cite{kipf2016semi}, which takes node features $\mathbf{H}^{(k)} \in \mathbb{R}^{N_G \times d}$ and the corresponding correlation matrix $\mathbf{M}$ as inputs and pass through a GCN layer to transform to $\mathbf{H}^{(k+1)} \in \mathbb{R}^{N_G \times d'}$. According to \cite{kipf2016semi}, every GCN layer can be represented as 
\begin{equation}
    \mathbf{H}^{(k+1)} = \text{LeakyReLU}(\hat{\mathbf{M}}\mathbf{H}^{(k)} \Theta^{(k)}),
\end{equation}
where $\Theta^{(k)} \in \mathbb{R}^{d\times d'}$ is a layer-specific trainable weight matrix and $\mathbf{\hat{M}}$ is the normalized version of correlation matrix $\mathbf{M}$. Formally, $\mathbf{\hat{M}}$ is defined as:
\begin{equation}
    \hat{\mathbf{M}} = (\mathbf{I} + \mathbf{D})^{-\frac{1}{2}}(\mathbf{M} + \mathbf{I})(\mathbf{I} + \mathbf{D})^{-\frac{1}{2}},
\end{equation}
where $\mathbf{D}$ is the diagonal degree matrix of $\mathbf{M}$, the identity matrix $\mathbf{I} \in \mathbb{R}^{N_G \times N_G}$ is added for forcing the self-loop in $\mathcal{G}$. We aim to learn a set of parameters $\Theta = \{\theta^{1}, \theta^{2}, ..., \theta^{k}\}$ that maps $\mathbf{X}$ to a set of inter-dependent classifier for person multi-attribute recognition.

\subsection{Training}
\label{sec:train}
\subsubsection{Attributes recognition}
In training process, we pass the feature map $\mathbf{F}$ to the global average pooling (GAP) layer to get the global feature $\mathbf{f}_{global}$ which is then passed through a BNNeck layer \cite{BoT} to get $\mathbf{f}_{bnn}$. For graph reasoning, the input node features $\mathbf{X}$ is fed to stacked of $k$-GCN layers and output node matrix $\mathbf{H}^{(k)} \in \mathbb{R}^{N_G \times D}$. We get a subset nodes $\mathcal{V}_a \subset \mathcal{V}$, where $v_a \in \mathcal{V}_a$ is considered as attribute nodes, and the corresponding output node features of $\mathcal{V}_a$ are stacked to form $\mathbf{W} \in \mathbb{R}^{N_A \times D}$ which is used to parameterized the multi-attributes classifier $\mathcal{C}_A$. The feature $\mathbf{f}_{bnn}$ is passed throught the classifier $\mathcal{C}_A$ to get the attribute prediction $\hat{\mathbf{y}}$
\begin{equation}
    \hat{\mathbf{y}} = \mathbf{W} \mathbf{f}_{bnn},
\end{equation}
Given $\hat{\mathbf{y}} \in \mathbb{R}^{N_A}$, prediction score of attribute $c$ is indicated as $\hat{y}_c$. We denote the ground-truth label of an image is $\mathbf{y} \in \{0,1\}^{N_A}$, where $y_c$ indicate whether attribute $c$ appears in the image or not. The cross-entropy loss function is adopted for multi-label recognition. Let $L_{a}^{(i)}$ be the attribute loss for probe image $I^{(i)}$, which is computed as
\begin{equation}
    \mathcal{L}_{a}^{(i)} = -\frac{1}{N_A}\sum\limits_{c=1}^{N_A}y_c^{(i)}\log(\sigma(\hat{y}_c^{(i)})) + (1 - y_c^{(i)})\log(1 - \sigma(\hat{y}_c^{(i)})),
    \label{eq:loss_a_one}
\end{equation}
where $\sigma(.)$ is the sigmoid function. The attribute loss for whole training set is computed as 
\begin{equation}
    \mathcal{L}_{a} = \frac{1}{N}\sum\limits_{i=1}^{N} \mathcal{L}_a^{(i)},
    \label{eq:loss_a}
\end{equation}
where $N$ is the number of samples in the training set. By optimizing the attribute recognition loss function $L_a$, the network implicitly models the correlation between person attributes and their associated visual body parts. However, the correlation between attributes and body parts is weakly linked here because it is not aware of the person identity information. To leverage all the information, we add the identity classification objective to the framework.

\subsubsection{Person identity classification} The purpose of identity classification is to discriminate the features of a person among all other identities. However, in the person ReID problem, the number of identities is significant. At the same time, the images of each identity are also varied due to the factor of variations (e.g., environment, camera viewpoint, pose, etc.). This intra-variation is a critical challenge for a person ReID system.

However, the attributes of one person do not change significantly when a probe image is captured from different cameras or poses. In GPS, the body part nodes are obtained from the feature map of the probe image; thus, they retain the given person's visual representation. The visual information is propagated to other nodes in the graph after passing through several GCN layers. Denoting the node features of graph $\mathcal{G}$ after passing through $l$ GCN layers is $\mathbf{H}^{(l)}$. We map the whole graph into a graph feature $\mathbf{f}_{graph} = \frac{1}{n} \sum_{i=1}^{n}\mathbf{h}^{(i)}$, where $\mathbf{h}^{(i)}$ is the representation of node $i$. By doing that, the graph features $\mathbf{f}_{graph}$ not only can represent the visual information as well as the semantic representation of the person (i.e., correlation of attributes) but also is robust to the addressed variations.

The graph features are then concatenated with person global features $\mathbf{f}_{bnn}$, the resulted features is used for identity classification. The identity prediction logits are computed as follow
\begin{equation}
    \mathbf{p} = \text{softmax}(\text{FC}(\mathbf{f}_{bnn} \odot \mathbf{f}_{graph})),
\end{equation}
where $\odot$ denotes the concatenate operation. Let $\mathbf{q}^{(i)}$ be the one-hot vector indicating the ground-truth identity and $\mathbf{p}^{(i)}$ is the identity prediction logits of image $i$. We use the Cross-entropy loss as folow
\begin{equation}
    \mathcal{L}_{id} = - \frac{1}{N}\sum\limits_{i=1}^{N} \mathbf{q}^{(i)}\log{\mathbf{p}^{(i)}}
    \label{eq:loss_id}
\end{equation}

\subsubsection{Multi-task loss}
The network in Figure \ref{fig:framework} is trained end-to-end using the following multi-task loss function
\begin{equation}
    \mathcal{L} = \alpha_1 \mathcal{L}_{id} + \alpha_2 \mathcal{L}_{triplet} + \alpha_3 \mathcal{L}_{center} + \alpha_4 \mathcal{L}_{a}
    \label{eq:final_loss}
\end{equation}
where $L_{id}$ is the identity loss (\ref{eq:loss_id}), $L_{triplet}$ is the triplet loss \cite{weinberger2006distance}, $L_{center}$ is the center loss \cite{centerloss}, and $L_{a}$ is attribute recognition loss (\ref{eq:loss_a}). Since attribute recognition and person identity classification use global features as input, the Triplet loss and Center loss are used to improve the representativeness of global features and generalize well to an unseen person in the test set.

\section{Experiments}
\label{sec:exp}
\subsection{Experimental Setup}

\begin{table*}
\centering
\caption{The contribution of losses to the performance of person ReID task on Market1501 \cite{market1501} dataset. Note that the experiments are conducted with ResNet-50 \cite{resnet} as backbone CNN network.}
\label{tab:contribution}
\begin{tabular}{p{1.2cm} | p{1.2cm} | p{1.2cm} | p{1.2cm}  p{1.2cm} | p{1.2cm}  p{1.2cm}}
\hline
\multirow{2}{*}{$\mathbf{\mathcal{L}_{id}}$} & \multirow{2}{*}{$\mathbf{\mathcal{L}_{triplet}}$} & \multirow{2}{*}{$\mathbf{\mathcal{L}_{center}}$} & \multicolumn{2}{c|}{\textbf{\textbf{without} $\mathbf{\mathcal{L}_{a}}$}} & \multicolumn{2}{c}{\textbf{with} $\mathbf{\mathcal{L}_{a}}$} \\ \cline{4-7}
                        &               &                   & mAP       & R-1       & mAP       & R-1           \\ \hline
\hline
\checkmark              &               &                   & 85.5      & 94.0      & 87.0      & 95.1        \\ \hline
\checkmark              & \checkmark    &                   & 87.1      & 94.7     & 87.6      & 95.2        \\ 
\hline
\checkmark              & \checkmark    & \checkmark        & 87.5      & 94.9     & 87.8      & 95.2        \\ \hline
\end{tabular}
\end{table*}

\begin{table*}
\centering
\caption{The transferable ability of our GPS evaluated on cross-dataset}
\label{tab:effect_cross}
\begin{tabular}{p{1.5cm} | p{1.0cm} p{1.5cm} p{1.5cm} | p{1.0cm} p{1.5cm} p{1.5cm}}
\hline
\multirow{2}{*}{Models} & \multicolumn{3}{c|}{Market-1501 $\rightarrow$ DukeMTMC-ReID} & \multicolumn{3}{c}{DukeMTMC-ReID $\rightarrow$ Market-1501} \\ \cline{2-7}
                &       & mAP       & R-1       &       & mAP       & R-1       \\ \hline \hline
BoT \cite{BoT}  &       & 14.6      & 27.6      &       & 21.6      & 48.6      \\ \hline
GPS (our)       &       & 21.9      & 37.0      &       & 24.7      & 52.1      \\ \hline
\end{tabular}
\end{table*}

\textbf{Implementation.} We integrate our GPS into the recent work BoT \cite{BoT} as the strong baseline. We employ the ResNet-50 \cite{resnet} pre-trained on ImageNet as the backbone network in all experiments. To enhance the discriminating power of the backbone, we integrate non-local attention (NLA) \cite{wang2018non} into each ResNet block. For each probe image, we resize them into 256 $\times$ 128 and pad the resized image 10 pixels with zero values. After that, we randomly crop them into a 256 $\times$ 128. For the data augmentation, similar to \cite{BoT,mancs,scal}, we use random horizontal flipping and erasing with the probability of 0.5 for both methods. Attribute labels are transformed into $N_A \times 300$ word embedding. Note that $N_A$ corresponds to the number of attributes of the dataset, i.e., 30 and 23 for Market1501 \cite{market1501} and DukeMTMC-ReID \cite{dukemtmc} dataset, respectively.

\textbf{Dataset.} To evaluate our proposed method, we conduct our experiments on two large-scale attribute person ReID datasets which are Market-1501 \cite{market1501} and DukeMTMC \cite{dukemtmc}. We follow the standard train/test split of each dataset in our experiments. 



\textbf{Evaluation.} To evaluate the person ReID performance of our GPS and to compare the results with the state-of-the-art methods, we report standard ReID metrics: Cumulative Matching Characteristic (CMC) (as R-1, R-5, and R-10) and mean Average Precision (mAP). 
Note that, as \cite{attreid}, we ignore the distractor and junks images which are not labelled attributes.

\subsection{GPS Analysis}

\textbf{Loss Contribution.}  In Table \ref{tab:contribution}, we show the contribution of each loss to the final performance on the Market1501 dataset. The person ID classification loss, triplet loss, center loss, and attribute recognition loss are denoted as $\mathbf{\mathcal{L}_{id}}$, $\mathbf{\mathcal{L}_{triplet}}$, $\mathbf{\mathcal{L}_{center}}$, and $\mathbf{\mathcal{L}_{a}}$, respectively. The performance is improved when we incorporate all losses to the framework, which justifies the effectiveness of our proposed method. By using only $\mathbf{\mathcal{L}_{id}}$, we still achieve comparative results with other mask-guided and attribute-based methods. While the triplet loss $\mathbf{\mathcal{L}_{triplet}}$ demonstrates its capability on improving the performance, the center loss $\mathbf{\mathcal{L}_{center}}$ shows a slight impact on the performance. Notably, the attribute loss $\mathbf{\mathcal{L}_{a}}$ shows stability when being incorporated with other loss functions.

\begin{table*}
\centering
\caption{The number of parameters of our GPS in comparision with the baseline BoT \cite{BoT} on Market1501 and DukeMTMC-ReID datasets using ResNet-50 \cite{resnet} as the backbone network. \#nParam indicates the number of parameters and 1K=1000.}
\label{tab:nParam}
\begin{tabular}{p{3.5cm} | p{3.0cm} | p{3.0cm}}
\hline
\multirow{2}{*}{Models}     &DukeMTMC-ReID &Market1501     \\
\cline{2-3}
                            &\#nParam (K)  &\#nParam (K)     \\
\hline
\hline
BoT \cite{BoT}              &25,668        &25,829       \\
GPS (our)                   &28,866        &28,715        \\
\hline
\end{tabular}
\end{table*}

\begin{table*}[!t]
	\caption{Comparison with state-of-the-art methods on Market-1501 \cite{market1501} and DukeMTMC-ReID \cite{dukemtmc} datasets. The cyan and yellow boxes are the best results corresponding to mask-guided/attribute-based and other approaches, respectively. Note that no post-processing is applied to our method.}
	\centering
	\begin{center}
		\begin{tabular}{p{2.5cm}| p{2.6cm} |p{0.7cm} |p{0.7cm} |p{0.7cm} |p{0.8cm}|p{0.7cm} |p{0.7cm} |p{0.7cm} |p{0.8cm}} 
			\hline
			\multirow{2}{*}{\textbf{Approach}}   & \multirow{2}{*}{\textbf{Method}}       & \multicolumn{4}{c|}{\textbf{Market1501}} & \multicolumn{4}{c}{\textbf{DukeMTMC-ReID}} \\ \cline{3-10}
			                            &                               &mAP   &R-1   &R-5   &R-10   &mAP    &R-1    &R-5    &R-10    \\
			\hline
			\hline
			\multirow{19}{*}{Others}     &SVDNet \cite{svdnet}           &62.1   &82.3   &92.3   &95.2   &56.8	&76.7	&86.4	&89.9       \\
			                            &TriNet \cite{triplet}          &69.1   &84.9   &94.2   &-      &-      &-      &-      &-          \\
			                            &AGW-att \cite{ye2020deep} &86.9    &94.9    &-  &-  &77.6   &87.5    &-  &-   \\
			                            &Pyramid \cite{pyramid}         &88.2   &95.7	&98.4	&99.0   &79.0	&89.0   &-      &-          \\
			                            &HPM \cite{hpm}                 &-      &-      &-      &-      &74.3	&86.6   &-      &-          \\
			                            &Auto-ReID \cite{auto-reid}	  	&85.1	&94.5	&-	    &-       &-      &-      &-      &-         \\
                                        &GCP \cite{gcp}                 &88.9	&95.2   &-      &-      &78.6	&89.7   &-      &-          \\
                                        &Mancs \cite{mancs}             &82.3	&93.1	&-      &-      &71.8	&84.9   &-      &-	        \\
                                        &SONA$^{2+3}$ \cite{sona}       &88.8	&95.6	&98.5	&99.2   &78.3	&89.4	&95.4	&96.6       \\
                                        &SCAL \cite{scal}      &\cellcolor[HTML]{F3F4BE}\textbf{89.3}	&95.8	&98.7   &-      &78.4	&88.6   &-      &-      \\
                                        &PN-GAN \cite{PN-GAN}           &72.6   &89.4	&-      &-      &53.2	&73.6   &-      &88.8       \\
                                        &FD-GAN \cite{FD-GAN}           &77.7   &90.5	&-      &-      &64.5	&80.0   &-      &-          \\
                                        &DG-Net \cite{dg-net}           &86.0   &94.8	&-      &-      &74.8	&86.6   &-      &-          \\
                                        &SGGNN \cite{sggnn}             &82.1	&92.3	&-	    &-      &68.2	&81.1   &-      &-          \\
                                        &st-ReID \cite{st-ReID}      &87.6	&\cellcolor[HTML]{F3F4BE}\textbf{98.1}	&\cellcolor[HTML]{F3F4BE}\textbf{99.3}	&\cellcolor[HTML]{F3F4BE}\textbf{99.6}  &\cellcolor[HTML]{F3F4BE}\textbf{83.9}	&\cellcolor[HTML]{F3F4BE}\textbf{94.4}	&\cellcolor[HTML]{F3F4BE}\textbf{97.4}	&\cellcolor[HTML]{F3F4BE}\textbf{98.2}   \\
                                        &CAMA \cite{cama}               &84.5	&94.7	&-      &-      &72.9	&85.8   &-      &-          \\
                                        &DSA \cite{dsa}                 &87.6	&95.7	&-      &-      &74.3	&86.2   &-      &-          \\
                                        &FPR \cite{fpr}                 &86.6	&95.4   &-      &-      &78.4	&88.6   &-      &-          \\		
                                        &SAN \cite{san}                 &88.0   &96.1   &-      &-      &75.5	&87.9   &-      &-          \\
			\hline
			\hline
			\multirow{12}{3.0cm}{Mask-guided \& \\ Attribute-based}       &MGCAM \cite{MGCAM}     &74.3	&83.8	&-  	&-      &-      &-      &-      &-          \\
                                        &SPReID \cite{SPReID}           &81.3	&92.5	&97.2	&98.1   &71.0	&84.4	&91.9	&93.7       \\
                                        &P$^2$-Net \cite{p2net}         &85.6	&\textbf{95.2}	&98.2	&99.1   &73.1	&86.5	&93.1	&95.0       \\ 
                                        &ACRN \cite{ACRN}               &62.6   &83.6   &92.6   &95.3   &52.0   &72.6   &84.8   &88.9       \\
                                    	&MLFN \cite{mlfn}               &74.3	&90.0	&-  	&-      &62.8	&81.0	&-      &-          \\
                                        &A$^3$M \cite{a3m}              &69.0	&86.5	&95.2	&97.0   &-      &-      &-      &-          \\
                                        &AANet \cite{aanet}             &82.5	&93.9	&-	    &98.6   &72.6	&86.4	&-      &-          \\
                                        &APR \cite{attreid}             &66.9	&87.0	&95.1	&96.4   &55.6	&73.9	&-      &-          \\ 
                                        &AFFNet \cite{affnet}           &81.7   &93.7   &-      &-      &70.7   &84.6   &-      &-          \\
                                        &PAAN \cite{PAAN}               &77.6   &92.4   &-      &-      &65.5   &82.6   &-      &-          \\
                                        &APDR \cite{apdr}               &80.1	&93.1	&97.2	&98.2   &69.7	&84.3	&92.4	&94.7       \\
                                        &BoT \cite{BoT}                 &85.9   &94.5   &-      &-      &76.4	&86.4   &-      &-          \\
                                        
                                        &GPS (ours)                  &\cellcolor[HTML]{B8F0E1}\textbf{87.8} &\cellcolor[HTML]{B8F0E1}\textbf{95.2} &\cellcolor[HTML]{B8F0E1}\textbf{98.4}  &\cellcolor[HTML]{B8F0E1}\textbf{99.1}  &\cellcolor[HTML]{B8F0E1}\textbf{78.7}  &\cellcolor[HTML]{B8F0E1}\textbf{88.2}  &\cellcolor[HTML]{B8F0E1}\textbf{95.2}  &\cellcolor[HTML]{B8F0E1}\textbf{96.7}\\
			
            \hline
		\end{tabular}
	\end{center}
	\label{tab:sota}
\end{table*}

\textbf{Model Interpretability.} In this section, we conduct cross-dataset experiments to evaluate the effectiveness of GPS. The model is trained on the source dataset and test directly on the target dataset without finetuning. As shown in Table \ref{tab:effect_cross}, our GPS archives a significant improvement over the Bag-of-Tricks baseline \cite{BoT}. This demonstrates the interpretability of our proposed method as well as confirms the effectiveness of learning the attributes for the person ReID task.

\textbf{Training Parameters.} We also provide the number of training parameters of our GPS and the baseline BoT \cite{BoT} in Table \ref{tab:nParam} to show the complexity of each method. Overall, our GPS slightly increases about 3M parameters in comparison with the baseline BoT while achieving much better performance.

\subsection{Comparison to the State of the Art}

\textbf{Mask-guided and Attribute-based Methods.} We compare our method (GPS) to the recent state-of-the-art that used body parts: MGCAM \cite{MGCAM}, SPReID \cite{SPReID}, P$^2$-Net \cite{p2net}. For attribute-based approach, we compare our results with ACRN \cite{ACRN}, MLFN \cite{mlfn}, A$^3$M \cite{a3m}, AANet \cite{aanet}, APR \cite{attreid}, AFFNet \cite{affnet}, PAAN \cite{PAAN}, and APDR \cite{apdr}. Among them, AANet \cite{aanet}, PAAN \cite{PAAN}, and APDR \cite{apdr} are works that use both attributes and body parts to enhance the performance of person ReID task. However, there is no work that leverage the relationship between attributes and body parts to extract person signature embedding as our proposed method. 

\textbf{Other Approaches.} We also compare our method with other ReID approaches, including global-based approach: SVDNet \cite{svdnet}, TriNet \cite{triplet}; stripes-based approach: Pyramid \cite{pyramid}, Auto-ReID \cite{auto-reid}, GCP \cite{MGCAM}; attention-based approach: Mancs \cite{mancs}, SONA$^{2+3}$ \cite{sona}, SCAL \cite{scal}; GAN-based approach: PN-GAN \cite{PN-GAN}, FD-GAN \cite{FD-GAN}, DG-Net \cite{dg-net}; graph-based approach: SGGNN \cite{sggnn}; spatial-temporal-based approach: st-ReID \cite{st-ReID}; other approaches: CAMA \cite{cama}, DSA \cite{dsa}, FPR \cite{fpr}, SAN \cite{san}. No post-processing such as re-ranking \cite{re-ranking} or multi-query fusion \cite{market1501} is applied to our method.

\textbf{GPS vs. Baseline}. The last two rows of Table \ref{tab:sota} show the result of our GPS when being integrated into the baseline BoT. The results clearly show that our GPS significantly improves the performance of BoT in both Market-1501 and DukeMTMC-ReID dataset. This demonstrates the effectiveness of our GPS and confirms the usefulness of learning the attributes in the ReID task.

\begin{figure*}
    \centering
    \includegraphics[width=0.8\textwidth, keepaspectratio=true]{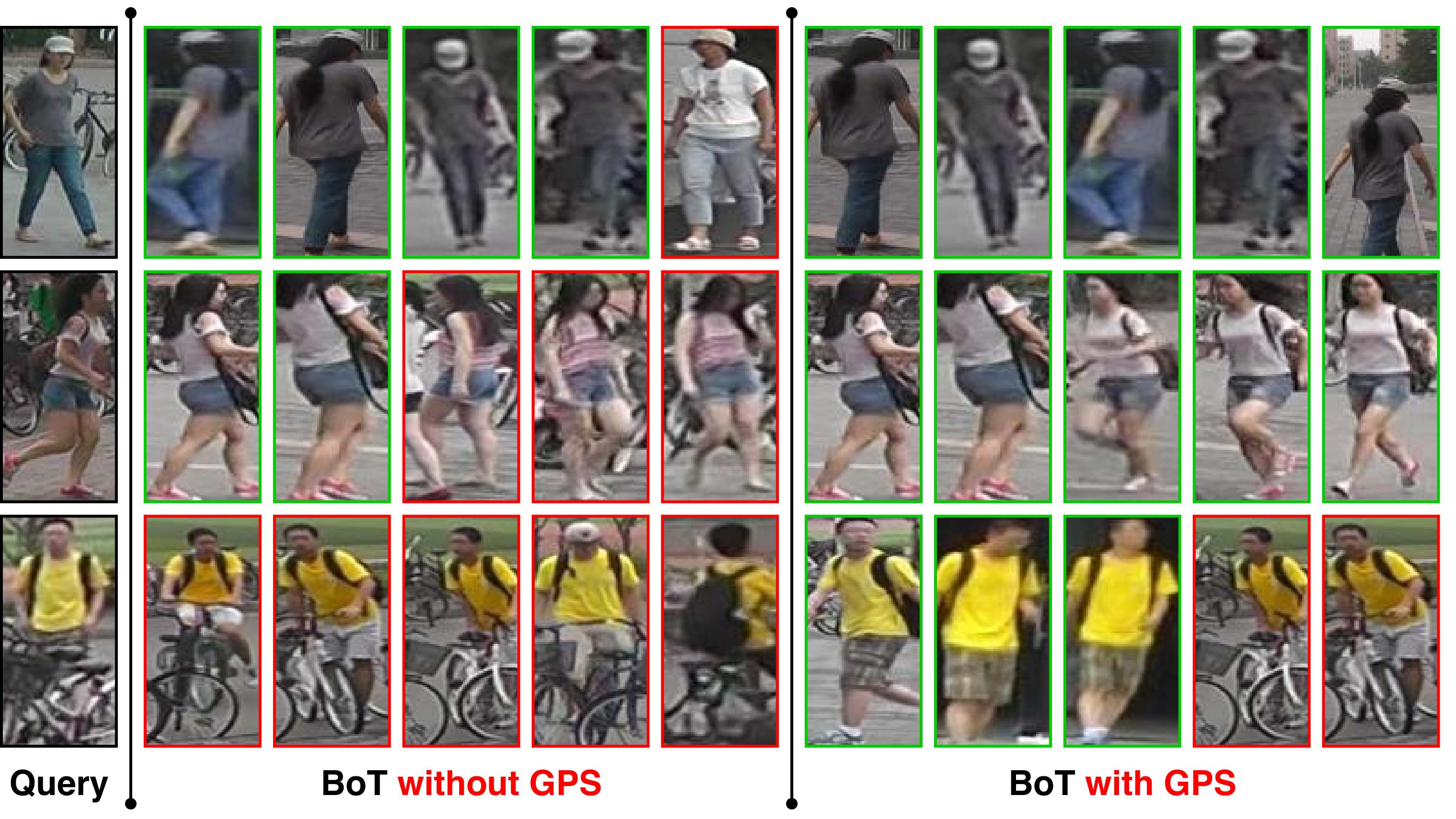}
    \caption{Top 5 retrieval results of some queries on Market-1501 dataset \cite{market1501}. Note that the green/red boxes denote true/false retrieval results, respectively.}
    \label{fig:visualization}
\end{figure*}

\textbf{Evaluation on Market-1501}. We evaluate our GPS with other methods on Market-1501 dataset in Table \ref{tab:sota}. The results show that our method outperforms the state-of-the-art attribute-based methods \cite{aanet} that use attribute and body part information in all evaluation metrics. Specifically, we outperforms AANet \cite{aanet} by 5.3\% and 1.3\% at mAP and R-1, respectively. Our GPS also outperforms the state-of-the-art mask-guided methods, and especially, we outperform P$^2$-Net \cite{p2net} by 2.2\% at mAP. At the same time, we also get comparative results when comparing with other recent ReID approaches.

\textbf{Evaluation on DukeMTMC-ReID}. Table \ref{tab:sota} also summaries the results of our GPS and other methods on DukeMTMC-ReID dataset. Our GPS significantly outperforms other attribute-based methods in all metrics. Specifically, our method outperforms the recent state-of-the-art attribute-based method AANet \cite{aanet} by 6.1\% at mAP and 1.8\% at R-1. In addition, we also outperforms ADPR \cite{apdr} by 9.0\%, 3.9\%, 2.8\%, 2.0\% at mAP, R-1, R-5, R-10, respectively. Moreover, our GPS outperforms the state-of-the-art mask-guided method P$^2$-Net \cite{p2net} by 4.9\%, 1.7\%, 2.1\%, 1.7\% at mAP, R-1, R-5, R-10, respectively. Besides, we also achieve comparative results with other ReID approaches.

\textbf{Attributed-based and Mask-guided vs. Other approaches.} From Table \ref{tab:sota}, we notice that although our GPS shows a definite improvement over mask-guided and attributed-based methods, it achieves competitive results with methods from other approaches and particularly being outperformed by st-ReID method \cite{st-ReID}. Note that the results of st-ReID also completely dominate all methods from all other approaches. The effectiveness of st-ReID comes from the fact that it also uses the spatial-temporal information (i.e., the spatial map of camera setting and temporal information from video timestamp) into the network. This extra information allows the network to encode the person identity from multiple viewpoints, which significantly reduces the effect of different poses, viewpoints, or ambiguity challenges. From experiments, we have observed that our GPS, as well as other attribute-based and mask-guided methods, suffers from the fact that the pretrained body part network cannot provide adequate segmentation masks, so the retrieval results are also affected.

We present some retrieval examples with five retrieved images for each query in Figure \ref{fig:visualization}. As in the visualization, our GPS obtained better retrieval results than the baseline. In the first row of Figure \ref{fig:visualization}, the baseline gets the false retrieval result at Rank-5 due to the similarity of gender, wearing a hat, etc., except the color of the clothes. By leveraging our GPS, the extracted features are more robust to attribute and body part information, then, lead to better retrieval results for ReID model. In the second row, the model with our GPS gives better results by extracting more information about the relationship between `backpack' attribute and this person identity, thereby eliminating false cases. We also show an example that our GPS does not yet produce entirely correct retrieval results in the third line of the Figure \ref{fig:visualization}. In this case, the lower body of the probe image is partly covered by the bicycle. Thus, the extracted features (i.e., the color of the pants) are not fully captured, which results in the feature misalignment between the probe image and retrieval results.

To conclude, our experiment results demonstrate that our GPS has successfully encoded two sources of local information (i.e., attributes and body parts) and global features, as well as modeling the correlations between them to create a visual signature of person identity. In the future, we would like to combine our approach and spatial-temporal information as in~\cite{st-ReID} to further improve the results.

\section{Conclusion}
\label{sec:conclusion}
This paper proposes Graph-based Person Signature (GPS) that effectively captures the dependencies of person attributes and body parts information. We utilize the GCN on the GPS to propagate the information among nodes in the graph and integrate the graph features into a novel multi-branch multi-task network. The experimental results on benchmark datasets confirm the effectiveness of our GPS and demonstrate that our GPS performs better than recent state-of-the-art attribute-based and mask-guided ReID methods.

{\small
\bibliographystyle{ieee_fullname}
\bibliography{egbib}
}

\end{document}